\documentclass[10pt,conference,a4paper]{IEEEtran}
\IEEEoverridecommandlockouts
\usepackage{cite}
\usepackage{amsmath,amssymb,amsfonts}
\usepackage{algorithmic}
\usepackage{graphicx}
\usepackage[hyphens]{url}
\usepackage{eurosym}
\usepackage{textcomp}
\usepackage{xcolor}
\usepackage{xspace}
\usepackage{subfig}
\usepackage{flushend}

\newcommand{\etal}{\textit{et~al}.\@\xspace}
\def\eg{\textit{e.g.}\@\xspace} 
\def\ie{\textit{i.e.}\@\xspace} 
 
\def\cf{\textit{cf.}\@\xspace} 
 
\def\wrt{w.r.t.\@\xspace}

\begin{document}

\title{Detection of Makeup Presentation Attacks based on Deep Face Representations}

\author{\IEEEauthorblockN{C. Rathgeb, P. Drozdowski, C. Busch}
\IEEEauthorblockA{\textit{da/sec -- Biometrics and Internet Security Research Group} \\
\textit{Hochschule Darmstadt, Germany}\\
\texttt{\{christian.rathgeb, pawel.drozdowski, christoph.busch\}@h-da.de}}
}

\maketitle

\begin{abstract}
Facial cosmetics have the ability to substantially alter the facial appearance, which can negatively affect the decisions of a face recognition. In addition, it was recently shown that the application of makeup can be abused to launch so-called makeup presentation attacks. In such attacks, the attacker might apply heavy makeup in order to achieve the facial appearance of a target subject for the purpose of impersonation. 

In this work, we assess the vulnerability of a COTS face recognition system to makeup presentation attacks employing the publicly available Makeup Induced Face Spoofing (MIFS) database. It is shown that makeup presentation attacks might seriously impact the security of the face recognition system. Further, we propose an attack detection scheme which distinguishes makeup presentation attacks from genuine authentication attempts by analysing differences in deep face representations obtained from potential makeup presentation attacks and corresponding target face images. The proposed detection system employs a machine learning-based classifier, which is trained with synthetically generated makeup presentation attacks utilizing a generative adversarial network for facial makeup transfer in conjunction with image warping. Experimental evaluations conducted using the MIFS database reveal a detection equal error rate of 0.7\% for the task of separating genuine authentication attempts from makeup presentation attacks.
\end{abstract}

\begin{IEEEkeywords}
Biometrics, face recognition, presentation attack detection, makeup attack detection, deep face representation
\end{IEEEkeywords}

\section{Introduction}
Presentation Attacks (PAs), a.k.a. \emph{spoofing attacks}, represent one of the most critical attack vectors against biometric recognition systems \cite{Marcel-HandbookPAD-ACVPR-2019}. This is particularly true for face recognition where numerous Presentation Attack Instruments (PAIs) can be employed by attackers, \eg face printouts or masks \cite{Galbally-Survey-Face-PAD-IEEEAccess-2014,Raghavendra-FacePAD-Survey-2017}. In the recent past, diverse countermeasures, \ie Presentation Attack Detection (PAD) methods, have been proposed for face recognition systems to prevent said attacks. For surveys on this topic the interested reader is referred to \cite{Galbally-Survey-Face-PAD-IEEEAccess-2014,Raghavendra-FacePAD-Survey-2017}. Published approaches can be categorised into software- and hardware-based PAD schemes where the latter make use of additional sensors, \eg depth or NIR capture devices \cite{Raghavendra-FacePAD-Survey-2017}. Even a PAI with high attack potential (\eg silicone masks) can be detected reliably with a spectral signature analysis of the skin which differentiates it from most artefacts \cite{Steiner-facePADswir-ICB-2016}.

\begin{figure}[!t]
\vspace{-0.0cm}
\centering
\subfloat[before]{\includegraphics[height=6.25cm]{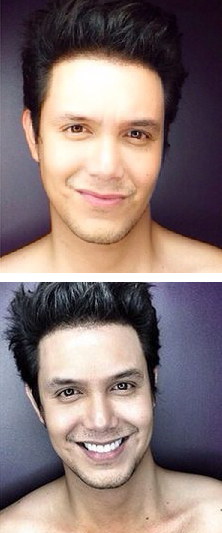}}\hfil
\subfloat[after]{\includegraphics[height=6.25cm]{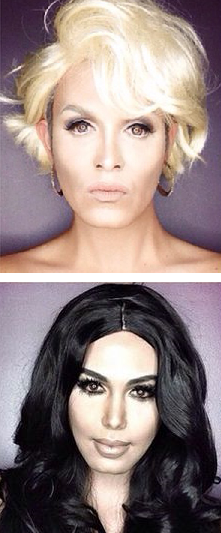}}\hfil
\subfloat[target]{\includegraphics[height=6.25cm]{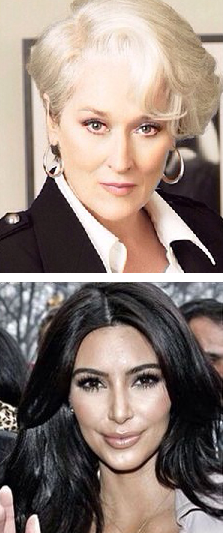}}\vspace{-0.0cm}
\caption{Web-collected examples of face images of the makeup artist Paolo Ballesteros (a) before and (b) after the application of makeup with the intention of obtaining the facial appearance of (c) target Hollywood stars. }\label{fig:example_makeup_artist}\vspace{-0.2cm}
\end{figure}

Recently, Chen~\etal \cite{Chen17a} showed that makeup can also be used to launch PAs. Makeup may substantially alter the perceived facial texture and shape which can pose a challenge to automated face recognition \cite{Dantcheva12a,Rathgeb-ImpactDetectionFacialBeautificationSurvey-ACCESS-2019}. When applied by skilled users or professional makeup artists, makeup can be abused with the aim of identity concealment or impersonation \cite{Chen17a}. In the latter case, makeup is applied in a way that the face of an attacker looks similar to that of a target subject, see figure~\ref{fig:example_makeup_artist}. Different makeup artists have showcased the possibility of transforming a face to that of a target subject through the mere application of makeup. Such Makeup PAs (M-PAs) pose a serious risk since these cannot be prevented by simply detecting makeup. More precisely, facial cosmetics are socially acceptable in many parts of the world and cultural communities. They have become a daily necessity for many women to improve facial aesthetics in a simple and cost-efficient manner \cite{Dantcheva12a}. This is evidenced by the huge and steadily growing market value of the facial cosmetics industry, \eg \euro77.6 billion in Europe in 2017 \cite{CosmeticsEurope} and \$63 billion in the US in 2016 \cite{MakeupFT}. This means, the mere use of makeup \emph{must not} be interpreted as a PA. It is important to note that this is not the case for other face PAIs species, which have been considered so far in the scientific literature, \eg face image printouts or three-dimensional (silicone) masks. Makeup might be used both in an innocent manner (bona fide subjects, who are interacting with the capture device in the fashion intended by the policy of the biometric system). However, it might as well be applied in a malicious manner (by subjects with the intent to impersonate an enrolled target). This clearly makes a reliable detection of M-PAs challenging. So far, only a few research efforts have been devoted to the topic of M-PAD, \eg in the ODIN research program \cite{Ericson-Odin-2018}.

In this work, we use standardised ISO/IEC methodology and metrics \cite{ISO-IEC-30107-3-PAD-metrics-170227} to evaluate the vulnerability of a Commercial Off-The-Shelf (COTS) face recognition system against M-PAs. Further, an image pair-based (\ie \emph{differential}) M-PAD system is introduced, which takes as input a potential M-PA and a target reference image. In this differential detection scenario, deep face representations are estimated from both face images employing state-of-the-art face recognition systems. Detection scores are obtained from machine learning-based classifiers analysing differences in deep face representations. Said classifiers are trained with a synthetic database of M-PAs (and bona fide authentication attempts) which are generated using a Generative Adversarial Network (GAN) for facial makeup transfer and image warping. In experiments, the publicly available MIFS dataset\footnote{Database available at \url{http://www.antitza.com/makeup-datasets.html}} is employed together with other publicly available face databases.

This paper is organised as follows: related works are discussed in section~\ref{sec:related}. The proposed M-PAD system is described in detail in section~\ref{sec:system}. The experimental setup is summarised in section~\ref{sec:setup} and experimental results are presented in section~\ref{sec:experiments}. Finally, conclusions are drawn in section~\ref{sec:conclusion}.

\section{Related Works}\label{sec:related}
Makeup induces non-permanent alterations with the ability to substantially alter the facial appearance. According to Dantcheva~\etal \cite{Dantcheva12a}, makeup can be applied mainly in three regions of the face, \ie eyes, lips, and skin. Prominent examples of makeup alterations include altering of the perceived contrast of the eyes, size of the mouth, as well as skin colour \cite{Dantcheva12a,Rathgeb-ImpactDetectionFacialBeautificationSurvey-ACCESS-2019}. Further, the application of makeup can be categorised \wrt intensity \cite{Dantcheva12a}, namely as light makeup (makeup cannot be easily perceived, since the applied colours correspond to natural skin, lip, and eye colours) and heavy makeup (makeup is clearly perceptible).

Dantcheva~\etal \cite{Dantcheva12a} were the first to systematically investigate the impact of facial makeup on face recognition systems. Performance degradations were observed in the case where either the reference or probe image had been altered by makeup. Similar studies confirming these findings were conducted in \cite{Eckert13a,wang2016recognizing}. Further, Ueda and Koyama \cite{Ueda10a} demonstrated that the use of heavy makeup significantly decreases humans' ability to recognise faces.

Towards achieving makeup-resilient face recognition, researchers have introduced different face feature extraction and comparison techniques. Several proposed approaches fused information obtained from multiple types of features, \eg \cite{moeini2014makeup,Kose15,Chen16a}. Additionally, different methods to detect makeup have been proposed, \eg \cite{Feng12a,Chen13a,Guo14a, wang2016face}. Such makeup detection schemes generally analyse facial colour, shape, and texture. In particular, skin features such as colour and smoothness were effectively extracted by applying suitable texture descriptors, \eg LBP and HOG, together with machine learning-based classifiers. If the application of facial makeup has been detected in a captured facial image, the face recognition system can react accordingly, \eg by applying the feature extraction with different parameters \cite{Rathgeb-ImpactDetectionFacialBeautificationSurvey-ACCESS-2019}.

\begin{figure}[!th]
\vspace{-0.0cm}
\centering
\begin{minipage}{0.32\linewidth}
\includegraphics[width=\linewidth]{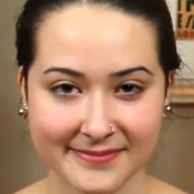}\newline\vspace{-0.25cm}
\subfloat[before]{\includegraphics[width=\linewidth]{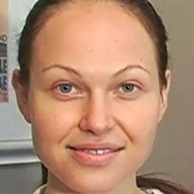}}
\end{minipage}\hspace{0.05cm}
\begin{minipage}{0.32\linewidth}
\includegraphics[width=\linewidth]{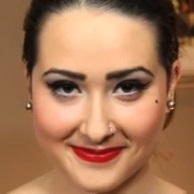}\newline\vspace{-0.25cm}
\subfloat[after]{\includegraphics[width=\linewidth]{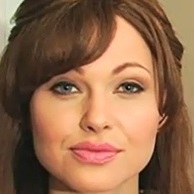}}
\end{minipage}\hspace{0.05cm}
\begin{minipage}{0.32\linewidth}
\includegraphics[width=\linewidth]{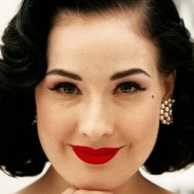}\newline\vspace{-0.25cm}
\subfloat[target]{\includegraphics[width=\linewidth]{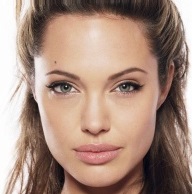}}
\end{minipage}\hspace{0.05cm}
\caption{Examples of cropped face images before and after the application of makeup intended to obtain the facial appearance of a target subject (images taken from the MIFS database).}\label{fig:example_makeup_target}\vspace{-0.0cm}
\end{figure}

Chen~\etal \cite{Chen17a} firstly investigated the potential of M-PAs with the aim of impersonation. To this end, the authors introduced the MIFS database, which was collected from YouTube makeup video tutorials containing face images of subjects before and after the application of makeup, as well as target victims. Example images of this database are shown in figure~\ref{fig:example_makeup_target}. Reported results suggested that different automated face recognition systems are vulnerable to M-PAs while the success chance of the attack is impacted by the appearance of the attacker and the target subject. Similarly, Zhu~\etal \cite{Zhu19} showed that the simulation of makeup in the digital domain can be used to launch adversarial attacks. More recently, Kotwal~\etal \cite{Kotwal} presented a deep learning-based M-PAD system which was designed to detect M-PAs aimed at identity concealment by emulating the effects of ageing. Interestingly, this scheme was also reported to achieve competitive detection performance on other databases in which makeup was applied for the purpose of facial beautification. This might suggest that the M-PAD system of \cite{Kotwal} detects different kinds of makeup. However, as mentioned before, the majority of subjects are not expected to wear makeup with the aim of identity concealment or impersonation but with the intent to beautify the overall facial impression. In a preliminary study, Rathgeb~\etal \cite{Rathgeb-Mpad-IWBF-2020} presented one of the first M-PAD systems in the scientific literature with the aim of detecting impersonation M-PAs.

Focusing on general face PAD, numerous software-based approaches have been presented in the last years \cite{Raghavendra-FacePAD-Survey-2017}. A comprehensive benchmark \cite{Boulkenafet17} which was conducted as part of a face PAD competition revealed that only some of the published approaches generalise against across PAIs and environmental conditions. In contrast, hardware-based approaches are expected to detect specific kinds of PAIs more reliably, but obviously require additional sensors \cite{Galbally-Survey-Face-PAD-IEEEAccess-2014}. 

\section{Makeup Presentation Attack Detection}\label{sec:system}
\begin{figure}[!th]
\vspace{-0.3cm}
\centering
\includegraphics[width=1.0\linewidth]{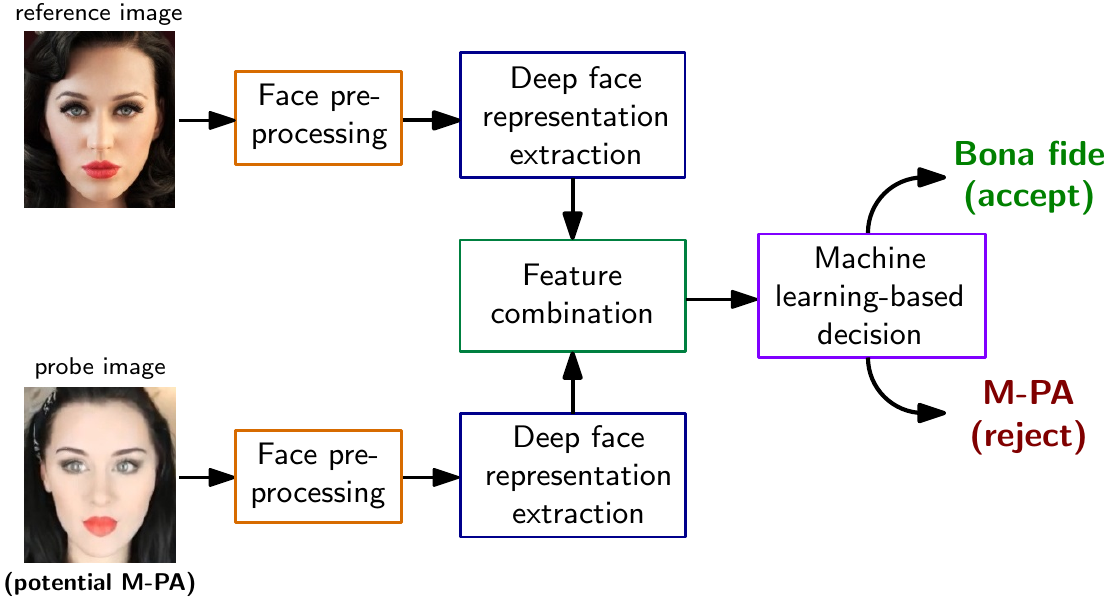}
\vspace{-0.4cm}
\caption{Overview of the proposed M-PAD system.}\label{fig:system}\vspace{-0.1cm}
\end{figure}

The following subsections describe the key components of the proposed M-PAD system, see figure~\ref{fig:system}. As mentioned earlier, M-PAs cannot be detected by simply detecting the presence of makeup, since makeup might as well be used by bona fide subjects. Therefore, a differential M-PAD system is designed which processes the stored reference image, in addition to the presented probe image. Differences between facial features extracted from a reference and a suspected probe image which indicate M-PAs are subsequently learned in a training stage employing a machine learning-based classifier. Similar differential attack detection systems have already been successfully proposed for face morphing \cite{Scherhag-FaceMorphingAttacks-TIFS-2020} and facial retouching \cite{Rathgeb-DifferentialDetectionRetouching-ACCESS-2020}.
The following subsections describe the employed extraction of deep face representations and the machine learning-based classification (section~\ref{sec:feature}) as well as the generation of synthetic M-PA training data (section~\ref{sec:training}).

\subsection{Feature Extraction and Classification}\label{sec:feature}
Given a pair consisting of a trusted reference image and a suspected probe image, faces are detected, normalised, and deep face representations are extracted from both images using a neural network of a state-of-the-art face recognition algorithm (see section~\ref{sec:dbs}). Deep face recognition systems leverage very large databases of face images to learn rich and compact representations of faces. It is expected that alterations induced by M-PAs will also be reflected in extracted deep face representations, outputs of the neural network on the lowest layer. Due to the high generalisation capabilities of deep face recognition systems with respect to variations in skin appearance, such changes might be more pronounced if the application of makeup changes the perceived facial shape. 

In principle, it would be possible to train a neural network from scratch or to apply transfer learning and re-train a pre-trained deep face recognition network to detect M-PAs. However, the high complexity of the model, represented by the large number of weights in the neural network, requires a large amount of training data. Even if only the lower layers are re-trained, the limited number of training images (and much lower number of subjects) in used databases can easily result in overfitting to the characteristics of the training set.
 
At classification, a pair of deep face representations extracted from a reference and probe face image are combined by estimating their difference vector. Specifically, an element-wise subtraction of feature vectors is performed. It is expected that differences in certain elements of difference vectors indicate M-PAs. In the training stage, difference vectors are extracted and a Support Vector Machine (SVM) with a Radial Basis Function (RBF) kernel is trained to distinguish bona fide authentication attempts and M-PAs. Alternatively, a concatenation of deep face representations extracted from reference and probe face images could be analysed. However, resulting feature vectors would potentially exhibit a length which prevent from an efficient classifier training. 

It is important to note that compared to the proposed differential M-PAD approach, a single image-based M-PAD system is not expected to reliably detect M-PAs \cite{Rathgeb-Mpad-IWBF-2020}. On the contrary, a single image-based M-PAD system which only analyses probe face images would most likely detect the mere application of makeup which does not indicate M-PAs per se, as explained in section~\ref{sec:related}.

\subsection{Training Data Generation}\label{sec:training}
Due to the fact that there exists no publicly available face database containing a sufficient amount of M-PAs to train a M-PAD system, we automatically generate a database of synthetic M-PAs. To this end, two face image transformations are applied, see figure~\ref{fig:train}:

\begin{figure}[!th]
\vspace{-0.3cm}
\centering
\includegraphics[width=1.0\linewidth]{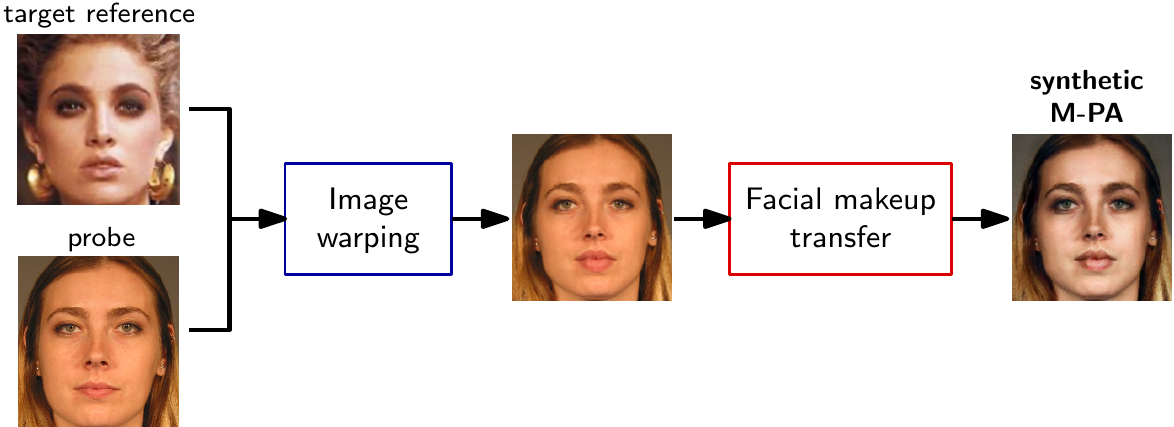}
\vspace{-0.3cm}
\caption{Processing steps of the generation of synthetic M-PAs.}\label{fig:train}\vspace{-0.0cm}
\end{figure}

\begin{enumerate}
\item \emph{Change of facial shape}: the shape of a face can be changed by heavily applying makeup, \eg slimming of contour and nose or enlargement of eyes, \cf figure~\ref{fig:example_makeup_artist}. To simulate said alterations with the aim of impersonating a target subject image warping \cite{Glasbey98} is applied. Specifically, facial landmarks of a target reference image and the probe image are extracted. Subsequently, image warping is applied to the probe face image \wrt the landmarks detected in the target reference image. The resulting probe face image will then exhibit the facial shape of the target reference image. This transformation is motivated by the fact that a skilled attacker, \eg makeup artist, would be able to change the appearance of his own facial shape through the application of makeup.
\item \emph{Change of facial texture}: the application of makeup can substantially alter the perceived texture of a face. In order to simulate textural changes induced by makeup with the aim of impersonating a target subject a GAN-based facial makeup transfer is employed (see section~\ref{sec:dbs}). GANs have enabled an automated transfer of full makeup styles, \eg \cite{Chang18,Li18}. Such transfer is motivated by the demand of users attempting to copy makeup styles of other individuals such as celebrities.
\end{enumerate} 

\begin{figure}[!th]
\vspace{-0.0cm}
\centering
\begin{minipage}{0.32\linewidth}
\includegraphics[width=\linewidth]{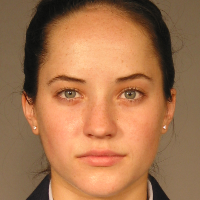}\newline\vspace{-0.25cm}
\includegraphics[width=\linewidth]{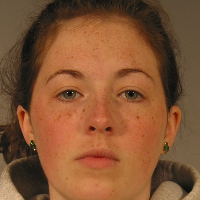}\newline\vspace{-0.25cm}
\subfloat[before]{\includegraphics[width=\linewidth]{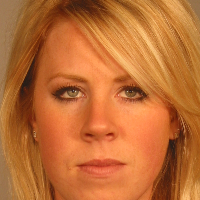}}
\end{minipage}\hspace{0.05cm}
\begin{minipage}{0.32\linewidth}
\includegraphics[width=\linewidth]{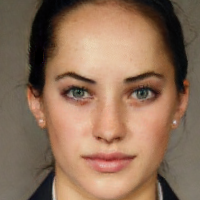}\newline\vspace{-0.25cm}
\includegraphics[width=\linewidth]{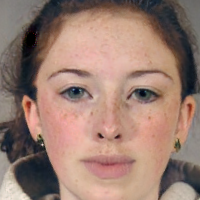}\newline\vspace{-0.25cm}
\subfloat[after]{\includegraphics[width=\linewidth]{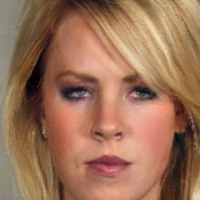}}
\end{minipage}\hspace{0.05cm}
\begin{minipage}{0.32\linewidth}
\includegraphics[width=\linewidth]{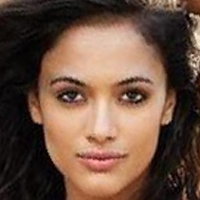}\newline\vspace{-0.25cm}
\includegraphics[width=\linewidth]{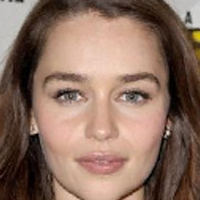}\newline\vspace{-0.25cm}
\subfloat[target]{\includegraphics[width=\linewidth]{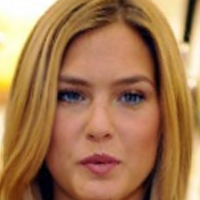}}
\end{minipage}\hspace{0.05cm}
\caption{Examples of cropped face images before and after the generation of synthetic M-PAs intended to obtain the facial appearance of a target subject.}\label{fig:synthetic}\vspace{-0.0cm}
\end{figure}

The aforementioned processing steps are applied to pairs of randomly chosen target reference images containing makeup and probe images of different subjects without makeup. For both types of images frontal pose, relatively neutral facial expression (\eg closed mouth), and sample image quality are automatically assured. Figure~\ref{fig:synthetic} depicts examples of resulting transformed probe images which represent synthetic M-PAs. Synthetic M-PAs are used in conjunction with unaltered pairs of face images of the same subject which represent bona fide authentication attempts.

The proposed synthetic generation of M-PAs could be adapted in several ways. On the one hand, the image warping process could be applied with a randomised intensity in order to simulate different skill levels of attackers. On the other hand, multiple  facial makeup transfer algorithms could be employed to improve robustness and avoid overfitting to potentially induced algorithm-specific artefacts. However, in the experimental setting used in this work, these adaptations did not reveal any improvements in terms of detection performance.

\section{Experimental Setup}\label{sec:setup}
The following subsections describe the software and databases (section~\ref{sec:dbs}), as well as evaluation methodology and metrics (section~\ref{sec:metrics}) used in the proposed M-PAD system and in experimental evaluations. 

\subsection{Software and Databases}\label{sec:dbs}
The \emph{dlib} algorithm \cite{King2009} is applied for face detection and facial landmark extraction. The detected eye coordinates are used for face alignment. Deep face representations are extracted using a re-implementation \cite{Facenetonline} of the well-known FaceNet algorithm \cite{Schroff-Facenet-CVPR-2015} and the open-source ArcFace system \cite{Deng19}. For both feature extractors, the resulting feature vectors consist of 512 floats. In addition, a COTS face recognition system is used in the vulnerability analysis. The use of the COTS face recognition system raises the practical relevance of the vulnerability analysis. While the COTS system is closed-source, it is assumed that it is based on deep learning as the vast majority of state-of-the-art face recognition systems. Therefore, it is only used in the vulnerability analysis, whereas open-source algorithms are used for the proposed M-PAD method.

During the generation of synthetic M-PAs, image warping is applied using \emph{OpenCV} with \emph{dlib} landmarks and a re-implementation \cite{BeautyGAN} of the \emph{BeautyGAN} algorithm of Li~\etal \cite{Li18} is used for facial makeup transfer. The \emph{scikit-learn} library \cite{scikit-learn} is used to train SVMs employing standard parameters. Trained SVMs generate a normalised attack detection score in the range $[0,1]$. 

\begin{table}[!th]
\centering
\caption{Overview of used databases.}\label{tab:dbs}\vspace{0.0cm}
\begin{footnotesize}
\begin{tabular}{|c|c|c|c|}
\hline
  \textbf{Purpose} &  \textbf{Bona Fide} &  \textbf{M-PAs} &  \textbf{Impostor} \\ \hline
 Vulnerability Assess. & MIFS, FRGCv2   &  MIFS  & FRGCv2\\
 M-PAD Training & FRGCv2  &  FRGCv2, CelebA  & -- \\
 M-PAD Testing &  MIFS, FERET &  MIFS  & -- \\
\hline
\end{tabular}
\end{footnotesize}\vspace{-0.0cm}
\end{table}

Table~\ref{tab:dbs} gives an overview of the used face image databases and their purposes. The MIFS face database is used to conduct the vulnerability assessment (see section~\ref{sec:vulnerability}). This database was introduced in \cite{Chen17a} and consists of 642 images of 117 subjects. For each subject three categories of images are available, \ie original face images, M-PAs, and face images of target subjects, see figure~\ref{fig:example_makeup_target}. The vulnerability assessment is done by performing comparisons between M-PAs and target images resulting in a total number of 428 M-PA attempts. It is important to note that bona fide image pairs of the MIFS face database exhibit almost no intra-class variation which is unlikely in a real-world scenario. Therefore, additional publicly available face image databases are used to obtain further bona fide authentication attempts which have realistic biometric variance, as it has also been done in \cite{Chen17a}. For the vulnerability analysis additional genuine score distributions are obtained from a subset of the FRGCv2 face database \cite{Phillips-2005}. Further, impostor score distributions are obtained from this database which contains 2,710 images of 533 subjects resulting in 3,298 genuine and 144,032 impostor comparisons. 

\begin{figure}[!t]
\vspace{-0.0cm}
\centering
\includegraphics[width=0.32\linewidth]{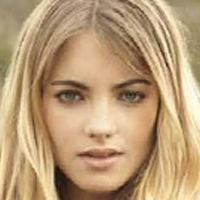}\hspace{0.05cm}
\includegraphics[width=0.32\linewidth]{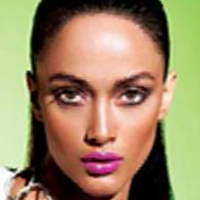}\hspace{0.05cm}
\includegraphics[width=0.32\linewidth]{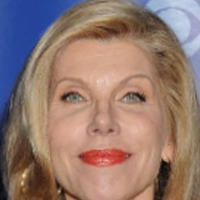}\\\vspace{0.15cm}
\includegraphics[width=0.32\linewidth]{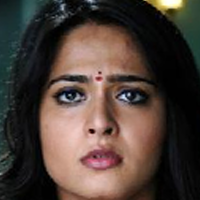}\hspace{0.05cm}
\includegraphics[width=0.32\linewidth]{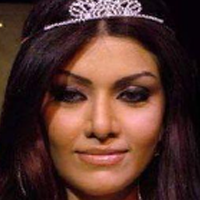}\hspace{0.05cm}
\includegraphics[width=0.32\linewidth]{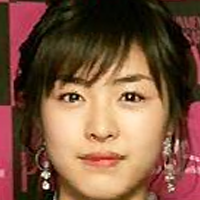}
\caption{Example face images of the subset of the CelebA database used for the training data generation for the proposed M-PAD system. }\label{fig:celebA}\vspace{-0.1cm}
\end{figure}

In the training stage of the proposed M-PAD system, a subset of the CelebA face database \cite{liu2015faceattributes} is used as target references. To obtain this subset, the CelebA face database has been filtered to only contain face images with heavy use of makeup, frontal pose, and closed mouth. Face sample quality assurance has been conducted using the FaceQNet algorithm \cite{Hernandez2019} resulting in a total number of 641 face images of different subjects. Example images of the resulting subset of the CelebA face database are depicted in figure~\ref{fig:celebA}. Images of the CelebA face database are randomly paired with face images of the FRGCv2 database to generate 3,290 synthetic M-PAs which are used together with the genuine authentication attempts of the FRGCv2 database to train the proposed M-PAD scheme.

In order to evaluate the detection performance of the M-PAD system, all M-PAs and bona fide comparisons of the MIFS database are used. Note that bona fide comparisons of the MIFS database include image pairs where makeup has been applied to one or both faces. A subset of the FERET face database \cite{Phillips:1998} is additionally employed to obtain 529 bona fide authentication attempts from different subjects. The use of the FERET face database is motivated by the fact that the FRGCv2 database has already been used in the training stage of the M-PAD system. 

\subsection{Evaluation Metrics}\label{sec:metrics}
Biometric performance is evaluated in terms of False Non-Match Rate (FNMR) and False Match Rate (FMR) \cite{ISO-PerformanceReporting-2006}. The performance of the M-PAD system is reported according to metrics defined in ISO/IEC 30107-3:2017 \cite{ISO-IEC-30107-3-PAD-metrics-170227} and its current revision. For the vulnerability assessment, the Impostor Attack Presentation Match Rate (IAPMR) \cite{ISO-IEC-30107-3-PAD-metrics-170227} defines the proportion of attack presentations using the same PAI species in which the target reference is matched. The Relative Impostor Attack Presentation Accept Rate (RIAPAR) establishes a relationship between IAPMR and 1$-$FNMR, \ie the biometric recognition performance of the attacked system, RIAPAR$=$1$+($IAPMR$-($1$-$FNMR$))$, as originally proposed by Scherhag \etal \cite{Scherhag-MorphingAttacks-MorphingTechniques-BIOSIG-2017}. The Attack Presentation Classification Error Rate (APCER) is defined as the proportion of attack presentations using the same PAI species incorrectly classified as bona fide presentations in a specific scenario. The Bona Fide Presentation Classification Error Rate (\mbox{BPCER}) is defined as the proportion of bona fide presentations incorrectly classified as PAs in a specific scenario. Further, as suggested in \cite{ISO-IEC-30107-3-PAD-metrics-170227} the BPCER10 and BPCER20 represent the operation points ensuring a security level of the system at an APCER of 10\% and 5\%, respectively. Additionally, the Detection Equal-Error Rate (D-EER) is reported.
 
\section{Experiments}\label{sec:experiments}
The following subsections summarise the conducted vulnerability assessment (section~\ref{sec:vulnerability}), considered baseline systems (section~\ref{sec:baseline}), and the obtained detection results of the M-PAD method (section~\ref{sec:detection}).

\subsection{Vulnerability Analysis}\label{sec:vulnerability} 
Table~\ref{tab:distribution_stats} and table~\ref{tab:iapmr} list statistics of score distributions and the vulnerability assessment, respectively. It can be observed that the IAPMRs and RIAPARs, \ie success chances, obtained by the original M-PAs are moderately high for practically relevant FMRs, \ie up to 17\% for an FMR of 1\%. This underlines the vulnerability of the face recognition system towards high quality M-PAs.

\begin{table}[!ht]
\centering
\caption{Descriptive statistics of score distributions.}\label{tab:distribution_stats}\vspace{0.0cm}
\begin{footnotesize}
\begin{tabular}{|r|c|c|c|c|}
\hline
    \textbf{Distribution} &   \textbf{Mean} &  \textbf{St. Dev.} &  \textbf{Minimum} &  \textbf{Maximum} \\
\hline
     Genuine &  0.945 &     0.033 &    0.021 &    0.996 \\
    Impostor &  0.057 &     0.064 &    0.000 &    0.905 \\
      Attack &  0.168 &     0.164 &    0.000 &    0.837 \\
\hline
\end{tabular}
\end{footnotesize}\vspace{-0.0cm}
\end{table}

\begin{table}[!ht]
\centering
\caption{Vulnerability in relation to biometric performance (in \%).}\label{tab:iapmr}\vspace{0.0cm}
\begin{footnotesize}
\begin{tabular}{|c|c|c|c|}
\hline
  \textbf{FMR} &  \textbf{FNMR} &  \textbf{IAPMR} &  \textbf{RIAPAR} \\ \hline
 0.001 &  6.274 &   0.000 &   6.274 \\
 0.010 &  0.083 &   2.103 &   2.186 \\
 0.100 &  0.028 &   6.308 &   6.336 \\
 1.000 &  0.028 &  17.056 &  17.084 \\
\hline
\end{tabular}
\end{footnotesize}\vspace{-0.0cm}
\end{table}

\subsection{Baseline Systems}\label{sec:baseline}

The proposed M-PAD system based on deep face representations is compared with different algotithms:
\begin{itemize}
	\item \textit{3D-Reconstruction (3D-R)}: the M-PAD approach of Rathgeb \etal \cite{Rathgeb-Mpad-IWBF-2020} extracts approximations of facial depth images from the reference and probe images. The distance between the depth values at all \emph{dlib} landmarks is computed using the MSE. The average pairwise MSE is returned as final M-PAD
score. The rationale of this scheme is that the approximated depth image of a probe image might significantly differ from that of the corresponding reference image in case of M-PAs.
	\item \textit{Facial Landmarks (FL)}: the aforementioned \emph{dlib} landmark detector \cite{King2009} is used to extract a total number of 68 two-dimensional facial landmarks from each reference and probe face image. Extracted landmarks describe the jawline, eyebrows, nose, eyes, and lips of a face.  Facial landmark positions are normalised according to eye coordinates. For the facial landmark-based feature vectors $x$- and $y$-coordinates are subtracted separately during feature combination, resulting in a difference vector of length 2$\times$68. Focusing on the task of M-PAD, positions of facial landmarks of the probe image might differ from that of the reference image if anatomical alterations induced by the M-PA do not precisely resemble that of the target subject. Similar schemes have been proposed for face image manipulation detection \cite{Rathgeb-DifferentialDetectionRetouching-ACCESS-2020}.
	\item \textit{Texture Descriptors (TD)}: at feature extraction, the aligned and cropped reference and probe images are converted to grayscale and divided into 4$\times$4 cells to retain local information. Local Binary Patterns (LBP) \cite{Ahonen04} are extracted from each cell of the pre-processed face images. LBP feature vectors are extracted employing a radius of one where eight neighboring pixel values are processed within  3$\times$3 pixel patches. For details on the extraction of LBP feature vectors, the reader is referred to \cite{Ahonen04}. Obtained feature values are aggregated in corresponding histograms. The final feature vector is formed as a concatenation of histograms extracted from each cell. LBP has been found to be a powerful feature for texture classification. It is expected that LBP-based feature vectors extracted from the reference and probe image clearly differ if the texture of the reference image differs from that of the probe image.  A similar scheme has been proposed in \cite{Scherhag-MorphingDetection-DAS-2018} for the purpose of  face image manipulation detection.
	\item \textit{Probe-only Deep Face Representation (P-DFR)}: lastly, deep face representations are extracted from the probe image only using the \emph{ArcFace} algorithm \cite{Deng19}. The extracted feature vector is then directly used to distinguish between M-PAs and bona fide authentications. The use of a probe-only M-PAD scheme should reveal whether is also possible to  detect M-PAs from single probe images. For this purpose, deep face representations are used since these represent rich textural as well as anatomical properties of face images. 
\end{itemize}
Apart from the approach in \cite{Rathgeb-Mpad-IWBF-2020}, all baseline systems utilise the same training set and SVM-based classifier as the proposed M-PAD system.

\begin{table}[!ht]
\centering
\caption{Error rates of the M-PAD baselines (in \%).}\label{tab:sota}\vspace{0.0cm}
\begin{footnotesize}
\begin{tabular}{|c|c|c|c|}
\hline
\textbf{Method} & \textbf{D-EER} & \textbf{BPCER10} & \textbf{BPCER20} \\
\hline
3D-R \cite{Rathgeb-Mpad-IWBF-2020} & 21.864 & 41.414 & 51.515 \\
FL \cite{Rathgeb-DifferentialDetectionRetouching-ACCESS-2020} & 16.428 & 23.727 & 29.818 \\
TD \cite{Scherhag-MorphingDetection-DAS-2018} & 33.803 & 93.569 & 97.917 \\
P-DFR & 38.785 & 83.652 & 90.068 \\
\hline
\end{tabular}
\end{footnotesize}\vspace{-0.1cm}
\end{table}

\subsection{Detection Performance}\label{sec:detection}
The performance rates of the baseline M-PAD systems are listed in table~\ref{tab:sota}. Best detection performance is achieved by the FL, associated to a moderate D-EER of approximately 16.5\%. Compared to the FL scheme, the 3D-R approach reveals slightly higher error rates resulting in a D-EER around 22\%. Generally, worse detection performance is obtained by the TD and P-DFR approaches. W.r.t. the TD, it can be concluded that a texture-based analysis is not suitable to distinguish between bona fide authentications and M-PAs. The P-DFR achieves a D-EER which is close to guessing. This underlines that a mere analysis of a suspected probe image might not be sufficient for M-PAD.

\begin{figure}[!th]

\centering
\subfloat[FaceNet]{\includegraphics[width=0.75\linewidth]{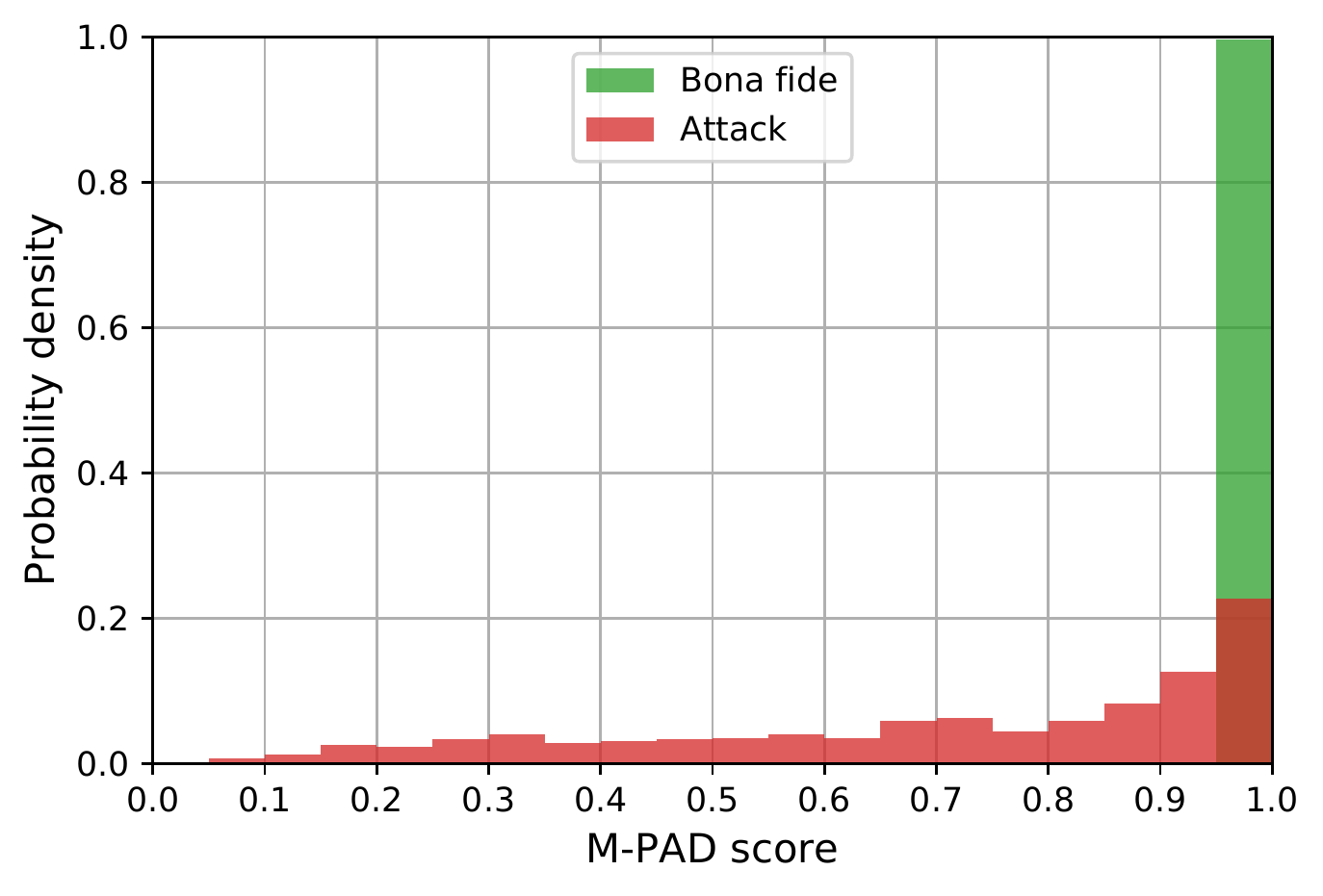}} \hfill
\subfloat[ArcFace]{\includegraphics[width=0.75\linewidth]{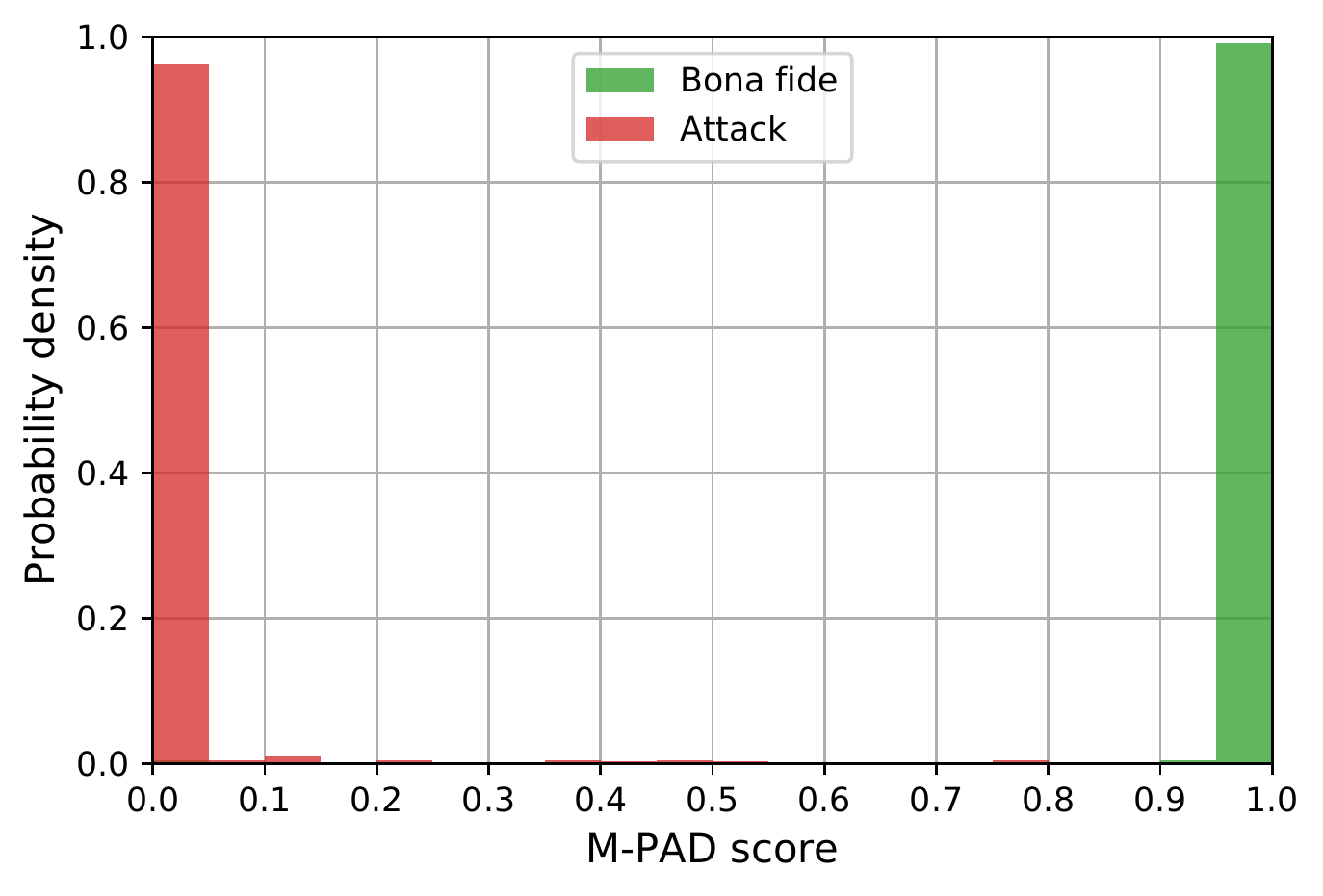}}\vspace{-0.0cm}
\caption{M-PAD score distributions for the proposed M-PAD.}\label{fig:scores}\vspace{-0.2cm}
\end{figure}

\begin{figure}[!ht]
\vspace{-0.0cm}
\centering
\includegraphics[height=5.5cm]{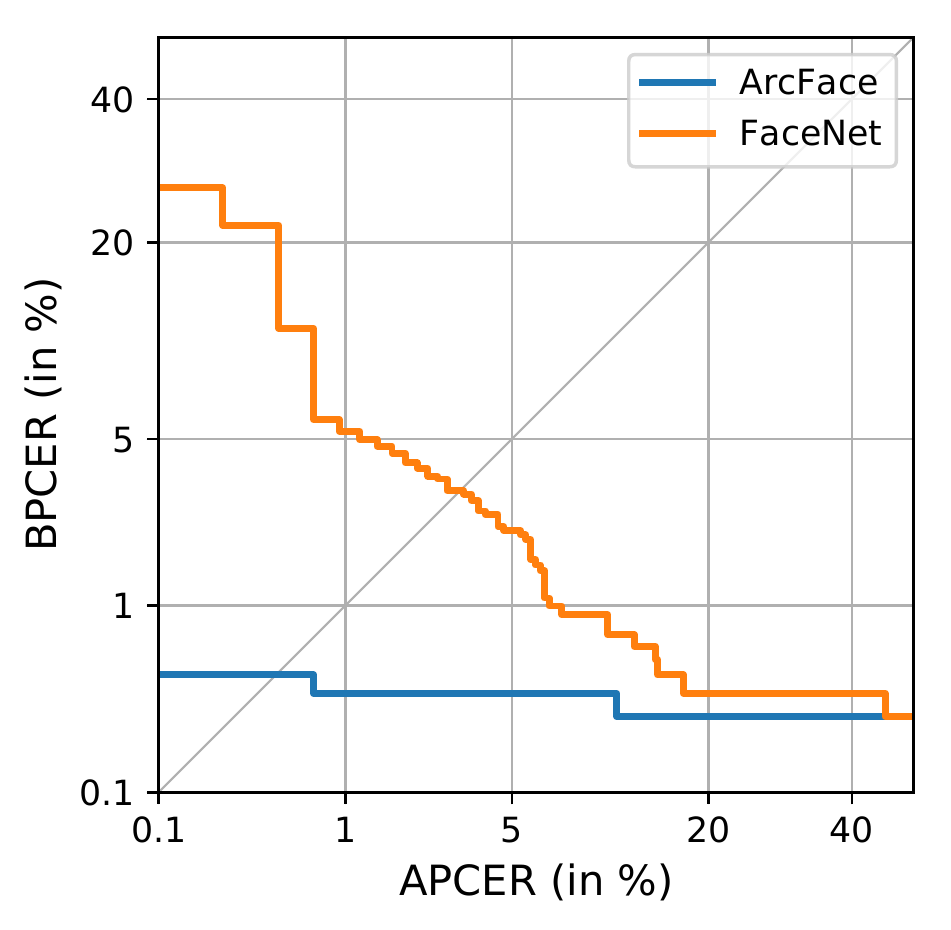}
\vspace{-0.2cm}
\caption{DET curves of proposed M-PAD.}\label{fig:det}\vspace{-0.1cm}
\end{figure}

Obtained detection accuracies of the M-PAD system for employing the different deep face representations are summarised in table~\ref{tab:detection}. Corresponding M-PAD score distributions are plotted in figure~\ref{fig:scores}. Corresponding DET curves are depicted in figure~\ref{fig:det}. It can be observed that the proposed M-PAD systems significantly outperform all considered baseline schemes. While the use of the FaceNet algorithm for the purpose of deep face representation extraction results in a D-EER of 3.271\%, the use of ArcFace achieves a D-EER of 0.701\% for the task of distinguishing bona fide authentication attempts from M-PAs.

\begin{table}[!ht]
\centering
\caption{Error rates of the proposed M-PAD (in \%).}\label{tab:detection}\vspace{0.0cm}
\begin{footnotesize}
\begin{tabular}{|c|c|c|c|}
\hline
     \textbf{Deep Face Representation} &   \textbf{D-EER} &  \textbf{BPCER10} &  \textbf{BPCER20} \\
\hline
   FaceNet & 3.271  &  0.904  &  2.169  \\
 ArcFace &  0.701 & 0.361   & 0.361  \\
\hline
\end{tabular}
\end{footnotesize}\vspace{-0.1cm}
\end{table}

\section{Conclusion and Future Work}\label{sec:conclusion}
We assessed the vulnerability of a COTS face recognition system against M-PAs. It was found that M-PAs of good quality, \ie ones which mimic the facial texture as well as the shape of an impersonated target subject, can pose a serious risk to the security of face recognition systems. In contrast, M-PAs based on a simple makeup style transfers have a rather low success rate.

In addition, we proposed a \emph{differential} M-PAD system which analyses differences in deep face representations extracted from a pair of reference and probe face images. Detection scores were obtained from SVM-based classifiers which have been trained to distinguish difference vectors from a training set of bona fide authentication attempts and synthetically generated M-PAs. In performance tests using the MIFS face database, the proposed M-PAD system was shown to achieve encouraging D-EERs of approximately 3.3\% and 0.7\% applying the FaceNet and ArcFace algorithm for the extraction of deep face representations, respectively. That is, the presented M-PAD scheme can effectively prevent M-PAs and hence improve the security of face recognition systems.

While the proposed M-PAD system makes use of a machine learning-based classifier which is trained with a few thousand synthetically generated images, an end-to-end deep learning-based M-PAD system is subject to future work. Such a system requires a huge amount of training data. Certainly, the presented generation of synthetic M-PAs would allow for the creation of a larger training database. However, the number of bona fide face images (in particular good quality reference images) is restricted by the employed databases. In order to avoid overfitting, large-scale face databases containing good quality images would be required to train an end-to-end deep learning-based M-PAD system.

\vspace{-0.0cm}
\section*{Acknowledgements}\label{sec:acknowledgements}
This research work has been funded by the German Federal Ministry of Education and Research and the Hessen State Ministry for Higher Education, Research and the Arts within their joint support of the National Research Center for Applied Cybersecurity ATHENE.

{\small
\bibliographystyle{IEEEtran}
\bibliography{references}

\begin{thebibliography}{10}
\providecommand{\url}[1]{#1}
\csname url@samestyle\endcsname
\providecommand{\newblock}{\relax}
\providecommand{\bibinfo}[2]{#2}
\providecommand{\BIBentrySTDinterwordspacing}{\spaceskip=0pt\relax}
\providecommand{\BIBentryALTinterwordstretchfactor}{4}
\providecommand{\BIBentryALTinterwordspacing}{\spaceskip=\fontdimen2\font plus
\BIBentryALTinterwordstretchfactor\fontdimen3\font minus
  \fontdimen4\font\relax}
\providecommand{\BIBforeignlanguage}[2]{{%
\expandafter\ifx\csname l@#1\endcsname\relax
\typeout{** WARNING: IEEEtran.bst: No hyphenation pattern has been}%
\typeout{** loaded for the language `#1'. Using the pattern for}%
\typeout{** the default language instead.}%
\else
\language=\csname l@#1\endcsname
\fi
#2}}
\providecommand{\BIBdecl}{\relax}
\BIBdecl

\bibitem{Marcel-HandbookPAD-ACVPR-2019}
S.~Marcel, M.~S. Nixon, J.~Fierrez, and N.~Evans, ``Handbook of biometric
  anti-spoofing: Presentation attack detection,'' 2019.

\bibitem{Galbally-Survey-Face-PAD-IEEEAccess-2014}
J.~Galbally, S.~Marcel, and J.~Fierrez, ``Biometric antispoofing methods: A
  survey in face recognition,'' \emph{IEEE Access}, vol.~2, pp. 1530--1552,
  December 2014.

\bibitem{Raghavendra-FacePAD-Survey-2017}
R.~Raghavendra and C.~Busch, ``Presentation attack detection methods for face
  recognition systems: {A} comprehensive survey,'' \emph{Computing Surveys
  ({CSUR})}, vol.~50, no.~1, pp. 1--37, March 2017.

\bibitem{Steiner-facePADswir-ICB-2016}
H.~Steiner, A.~Kolb, and N.~Jung, ``Reliable face anti-spoofing using
  multispectral {SWIR} imaging,'' in \emph{International Conference on
  Biometrics ({ICB})}.\hskip 1em plus 0.5em minus 0.4em\relax IEEE, June 2016,
  pp. 1--8.

\bibitem{Chen17a}
C.~{Chen}, A.~{Dantcheva}, T.~{Swearingen}, and A.~{Ross}, ``Spoofing faces
  using makeup: An investigative study,'' in \emph{International Conference on
  Identity, Security and Behavior Analysis ({ISBA})}.\hskip 1em plus 0.5em
  minus 0.4em\relax IEEE, February 2017, pp. 1--8.

\bibitem{Dantcheva12a}
A.~{Dantcheva}, C.~{Chen}, and A.~{Ross}, ``Can facial cosmetics affect the
  matching accuracy of face recognition systems?'' in \emph{International
  Conference on Biometrics: Theory, Applications and Systems ({BTAS})}.\hskip
  1em plus 0.5em minus 0.4em\relax IEEE, September 2012, pp. 391--398.

\bibitem{Rathgeb-ImpactDetectionFacialBeautificationSurvey-ACCESS-2019}
C.~Rathgeb, A.~Dantcheva, and C.~Busch, ``Impact and detection of facial
  beautification in face recognition: An overview,'' \emph{IEEE Access},
  vol.~7, pp. 152\,667--152\,678, October 2019.

\bibitem{CosmeticsEurope}
{Cosmetics Europe}, ``Socio-economic contribution of the {E}uropean cosmetics
  industry,''
  \url{https://www.ft.com/content/4721ed6a-f797-11e5-96db-fc683b5e52db}, May
  2018, last accessed: \today.

\bibitem{MakeupFT}
L.~Whipp, ``Changing face of cosmetics alters \$63bn {US} beauty market,''
  \url{https://www.ft.com/content/4721ed6a-f797-11e5-96db-fc683b5e52db}, April
  2016, last accessed: \today.

\bibitem{Ericson-Odin-2018}
L.~Ericson, ``Overview of the odin program on presentation attack detection,
  {International Face Performance Conference (IFPC)},'' 2018.

\bibitem{ISO-IEC-30107-3-PAD-metrics-170227}
{ISO/IEC JTC 1/SC 37 Biometrics}, \emph{{ISO/IEC} 30107-3. Information
  Technology -- Biometric presentation attack detection -- Part~3: Testing and
  Reporting}, September 2017.

\bibitem{Eckert13a}
M.~{Eckert}, N.~{Kose}, and J.-L. {Dugelay}, ``Facial cosmetics database and
  impact analysis on automatic face recognition,'' in \emph{International
  Workshop on Multimedia Signal Processing ({MMSP})}.\hskip 1em plus 0.5em
  minus 0.4em\relax IEEE, September 2013, pp. 434--439.

\bibitem{wang2016recognizing}
T.~Y. Wang and A.~Kumar, ``Recognizing human faces under disguise and makeup,''
  in \emph{International Conference on Identity, Security and Behavior Analysis
  ({ISBA})}.\hskip 1em plus 0.5em minus 0.4em\relax IEEE, February 2016, pp.
  1--7.

\bibitem{Ueda10a}
S.~Ueda and T.~Koyama, ``Influence of make-up on facial recognition,''
  \emph{Perception}, vol.~39, no.~2, pp. 260--264, February 2010.

\bibitem{moeini2014makeup}
A.~Moeini, H.~Moeini, F.~Ayatollahi, and K.~Faez, ``Makeup-invariant face
  recognition by {3D} face: Modeling and dual-tree complex wavelet transform
  from women's {2D} real-world images,'' in \emph{International Conference on
  Pattern Recognition ({ICPR})}.\hskip 1em plus 0.5em minus 0.4em\relax IEEE,
  August 2014, pp. 1710--1715.

\bibitem{Kose15}
N.~{Kose}, L.~{Apvrille}, and J.-L. {Dugelay}, ``Facial makeup detection
  technique based on texture and shape analysis,'' in \emph{International
  Conference and Workshops on Automatic Face and Gesture Recognition ({FG})},
  vol.~1.\hskip 1em plus 0.5em minus 0.4em\relax IEEE, May 2015, pp. 1--7.

\bibitem{Chen16a}
C.~Chen, A.~Dantcheva, and A.~Ross, ``An ensemble of patch-based subspaces for
  makeup-robust face recognition,'' \emph{Information Fusion}, vol.~32, no.~B,
  pp. 80--92, November 2016.

\bibitem{Feng12a}
R.~{Feng} and B.~{Prabhakaran}, ``Quantifying the makeup effect in female faces
  and its applications for age estimation,'' in \emph{International Symposium
  on Multimedia ({ISM})}.\hskip 1em plus 0.5em minus 0.4em\relax IEEE, December
  2012, pp. 108--115.

\bibitem{Chen13a}
C.~{Chen}, A.~{Dantcheva}, and A.~{Ross}, ``Automatic facial makeup detection
  with application in face recognition,'' in \emph{International Conference on
  Biometrics ({ICB})}.\hskip 1em plus 0.5em minus 0.4em\relax IEEE, June 2013,
  pp. 1--8.

\bibitem{Guo14a}
G.~{Guo}, L.~{Wen}, and S.~{Yan}, ``Face authentication with makeup changes,''
  \emph{Transactions on Circuits and Systems for Video Technology ({TCSVT})},
  vol.~24, no.~5, pp. 814--825, August 2014.

\bibitem{wang2016face}
S.~Wang and Y.~Fu, ``Face behind makeup,'' in \emph{Conference on Artificial
  Intelligence}.\hskip 1em plus 0.5em minus 0.4em\relax AAAI, February 2016,
  pp. 58--64.

\bibitem{Zhu19}
Z.~{Zhu}, Y.~{Lu}, and C.~{Chiang}, ``Generating adversarial examples by makeup
  attacks on face recognition,'' in \emph{International Conference on Image
  Processing ({ICIP})}.\hskip 1em plus 0.5em minus 0.4em\relax IEEE, September
  2019, pp. 2516--2520.

\bibitem{Kotwal}
K.~{Kotwal}, Z.~{Mostaani}, and S.~{Marcel}, ``Detection of age-induced makeup
  attacks on face recognition systems using multi-layer deep features,''
  \emph{Transactions on Biometrics, Behavior, and Identity Science ({TBIOM})},
  pp. 1--11, October 2019.

\bibitem{Rathgeb-Mpad-IWBF-2020}
C.~Rathgeb, P.~Drozdowski, D.~Fischer, and C.~Busch, ``Vulnerability assessment
  and detection of makeup presentation attacks,'' in \emph{International
  Workshop on Biometrics and Forensics ({IWBF})}.\hskip 1em plus 0.5em minus
  0.4em\relax IEEE, April 2020.

\bibitem{Boulkenafet17}
Z.~{Boulkenafet}, J.~{Komulainen}, Z.~{Akhtar}, A.~{Benlamoudi}, D.~{Samai}
  \emph{et~al.}, ``A competition on generalized software-based face
  presentation attack detection in mobile scenarios,'' in \emph{International
  Joint Conference on Biometrics ({IJCB})}.\hskip 1em plus 0.5em minus
  0.4em\relax IEEE, October 2017, pp. 688--696.

\bibitem{Scherhag-FaceMorphingAttacks-TIFS-2020}
U.~Scherhag, C.~Rathgeb, J.~Merkle, and C.~Busch, ``Deep face representations
  for differential morphing attack detection,'' \emph{{IEEE} Transactions on
  Information Forensics and Security}, 2020.

\bibitem{Rathgeb-DifferentialDetectionRetouching-ACCESS-2020}
C.~Rathgeb, C.-I. Satnoianu, N.~E. Haryanto, K.~Bernardo, and C.~Busch,
  ``Differential detection of facial retouching: A multi-biometric approach,''
  \emph{IEEE Access}, vol.~8, pp. 106\,373--106\,385, June 2020.

\bibitem{Glasbey98}
C.~A. Glasbey and K.~V. Mardia, ``A review of image-warping methods,''
  \emph{Journal of Applied Statistics}, vol.~25, no.~2, pp. 155--171, April
  1998.

\bibitem{Chang18}
H.~Chang, J.~Lu, F.~Yu, and A.~Finkelstein, ``{PairedCycleGAN}: Asymmetric
  style transfer for applying and removing makeup,'' in \emph{International
  Conference on Computer Vision and Pattern Recognition ({CVPR})}.\hskip 1em
  plus 0.5em minus 0.4em\relax IEEE, June 2018.

\bibitem{Li18}
T.~Li, R.~Qian, C.~Dong, S.~Liu, Q.~Yan, W.~Zhu, and L.~Lin, ``{BeautyGAN}:
  Instance-level facial makeup transfer with deep generative adversarial
  network,'' in \emph{International Conference on Multimedia ({MM})}.\hskip 1em
  plus 0.5em minus 0.4em\relax ACM, October 2018, pp. 645--653.

\bibitem{King2009}
D.~E. King, ``{Dlib-ml}: A machine learning toolkit,'' \emph{Journal of Machine
  Learning Research ({JMLR})}, vol.~10, pp. 1755--1758, December 2009.

\bibitem{Facenetonline}
D.~Sandberg, ``{Face recognition using Tensorflow},''
  \url{https://github.com/davidsandberg/facenet}, last accessed: \today.

\bibitem{Schroff-Facenet-CVPR-2015}
F.~Schroff, D.~Kalenichenko, and J.~Philbin, ``{FaceNet}: A unified embedding
  for face recognition and clustering,'' in \emph{Conference on Computer Vision
  and Pattern Recognition ({CVPR})}.\hskip 1em plus 0.5em minus 0.4em\relax
  IEEE, June 2015, pp. 815--823.

\bibitem{Deng19}
J.~Deng, J.~Guo, N.~Xue, and S.~Zafeiriou, ``Arcface: Additive angular margin
  loss for deep face recognition,'' in \emph{Conference on Computer Vision and
  Pattern Recognition ({CVPR})}.\hskip 1em plus 0.5em minus 0.4em\relax IEEE,
  June 2019, pp. 4690--4699.

\bibitem{BeautyGAN}
H.~Zhang, ``{BeautyGAN}: Instance-level facial makeup transfer with deep
  generative adversarial network,'' \url{https://github.com/Honlan/BeautyGAN},
  last accessed: \today.

\bibitem{scikit-learn}
F.~Pedregosa \emph{et~al.}, ``Scikit-learn: Machine learning in {P}ython,''
  \emph{Journal of Machine Learning Research ({JMLR})}, vol.~12, pp.
  2825--2830, October 2011.

\bibitem{Phillips-2005}
P.~J. Phillips, P.~J. Flynn, T.~Scruggs, K.~W. Bowyer, J.~Chang, K.~Hoffman,
  J.~Marques, J.~Min, and W.~Worek, ``Overview of the face recognition grand
  challenge,'' in \emph{Conference on Computer Vision and Pattern Recognition
  ({CVPR})}, vol.~1.\hskip 1em plus 0.5em minus 0.4em\relax IEEE, June 2005,
  pp. 947--954.

\bibitem{liu2015faceattributes}
Z.~Liu, P.~Luo, X.~Wang, and X.~Tang, ``Deep learning face attributes in the
  wild,'' in \emph{International Conference on Computer Vision ({ICCV})}.\hskip
  1em plus 0.5em minus 0.4em\relax IEEE, December 2015, pp. 3730--3738.

\bibitem{Hernandez2019}
J.~Hernandez{-}Ortega, J.~Galbally, J.~Fi{\'{e}}rrez, R.~Haraksim, and
  L.~Beslay, ``{FaceQnet}: Quality assessment for face recognition based on
  deep learning,'' in \emph{International Conference on Biometrics
  ({ICB})}.\hskip 1em plus 0.5em minus 0.4em\relax IEEE, June 2019.

\bibitem{Phillips:1998}
J.~Phillips, H.~Wechsler, J.~Huang, and P.~Rauss, ``The {FERET} database and
  evaluation procedure for face recognition algorithms,'' \emph{Image and
  Vision Computing Journal ({IMAVIS})}, vol.~16, no.~5, pp. 295--306, April
  1998.

\bibitem{ISO-PerformanceReporting-2006}
{ISO/IEC JTC1 SC37 Biometrics}, \emph{{ISO/IEC} 19795-1:2006. Information
  Technology -- Biometric Performance Testing and Reporting -- Part~1:
  Principles and Framework}, April 2006.

\bibitem{Scherhag-MorphingAttacks-MorphingTechniques-BIOSIG-2017}
U.~Scherhag, A.~Nautsch, C.~Rathgeb, M.~Gomez-Barrero, R.~N.~J. Veldhuis
  \emph{et~al.}, ``Biometric systems under morphing attacks: Assessment of
  morphing techniques and vulnerability reporting,'' in \emph{International
  Conference of the Biometrics Special Interest Group ({BIOSIG})}.\hskip 1em
  plus 0.5em minus 0.4em\relax IEEE, September 2017, pp. 1--7.

\bibitem{Ahonen04}
T.~Ahonen, A.~Hadid, and M.~Pietik{\"a}inen, ``Face recognition with local
  binary patterns,'' in \emph{European Conf. on Computer Vision
  (ECCV'04)}.\hskip 1em plus 0.5em minus 0.4em\relax Berlin, Heidelberg:
  Springer Berlin Heidelberg, 2004, pp. 469--481.

\bibitem{Scherhag-MorphingDetection-DAS-2018}
U.~Scherhag, C.~Rathgeb, and C.~Busch, ``Towards detection of morphed face
  images in electronic travel documents,'' in \emph{13th IAPR Workshop on
  Document Analysis Systems (DAS)}, 2018, pp. 1--6.

\end{thebibliography}
}

\end{document}